\documentclass[runningheads]{llncs}
\usepackage[T1]{fontenc}
\usepackage{graphicx}
\usepackage{hyperref}
\usepackage{amsmath}
\usepackage{booktabs}
\usepackage{multirow}
\usepackage{tabularx}
\usepackage{textcomp}
\usepackage{cleveref}
\usepackage[dvipsnames]{xcolor}

\usepackage{xcolor}
\definecolor{hardgreen}{HTML}{97d077}
\definecolor{softgreen}{HTML}{D5E8D4}
\definecolor{softyellow}{HTML}{fff2cc}
\definecolor{hardred}{HTML}{ff9999}
\definecolor{softred}{HTML}{f8cecc}

\begin{document}
\title{Adaptive Code Generation for Controlling Robots}
%
%
\author{
Justus Flerlage\inst{1} \and
Thorsten Wittkopp\inst{1,2} \and
Alexander Acker\inst{2} \and
Odej Kao\inst{1}
}

\authorrunning{J. Flerlage et al.}

\institute{
Technische Universität Berlin, Germany \\
\email{\{j.flerlage, t.wittkopp, odej.kao\}@tu-berlin.de} \and
logsight.ai, Berlin, Germany \\
\email{\{thorsten.wittkopp, alexander.acker\}@logsight.ai}
}

\maketitle              
\begin{abstract}
Deploying robots as Complex Adaptive Systems (CAS) in unknown and dynamic environments necessitates a transition from rigid command libraries toward intention-based autonomy, as natural language represents the only medium capable of articulating complex goals beyond the capacity of finite instruction sets.
While Large Language Models (LLMs) offer a path toward natural language goal description, their integration introduces significant challenges: the formalization gap between imprecise intentions and executable actions, the taxonomy gap induced by unpredictable environments, and the challenge of maintaining temporal state and progress awareness.
This work introduces an architectural framework that enables robotic control by leveraging generative AI.
The system follows a dual-AI design: an LLM translates high-level intentions into executable program code restricted to a formal robotic library and constrained by verifiable syntax, while a Vision-Language Model (VLM) provides semantic grounding via a distillation process. 
To ensure robustness, the framework incorporates environment-driven replanning triggers based on geometric and semantic thresholds, complemented by continuous runtime monitoring and an adaptive planning loop. 
Benchmarked across frontier models, our framework architecture demonstrates that grounding generative AI in a reactive, constrained loop enables robust fulfillment of complex intentions in dynamic and unknown environments.

\end{abstract}

\section{Introduction}
CAS are designed to operate under uncertainty, partial knowledge, and continuous environmental change. 
Rather than relying on a single centralized controller with a complete model of the world, CAS coordinates heterogeneous components whose collective behavior emerges through feedback, adaptation, and interaction with a dynamic environment~\cite{wirsing2023rigorous,alrahman2017abc}. 
Robotic systems operating outside static industrial settings constitute a particularly challenging instance of this paradigm. 
They must interpret high-level goals, perceive incomplete and changing environments, select appropriate actions, and revise their behavior when assumptions about the world become invalid.

However, conventional robotic systems are typically designed for static and well-defined environments and tasks. 
The prevailing paradigms of conventional robotics have historically centered on low-level control, symbolic planning, and rule-based systems~\cite{jiang2019task,sridharan2019reba}.
Within industrial environments, such as automotive assembly lines, robots execute predetermined operations based on a closed-world assumption, utilizing deterministic algorithms, fixed rule sets, and task models that are entirely specified in advance.
Even in domains where Artificial Intelligence (AI) appears to provide greater flexibility, such as autonomous driving or industrial manipulation, system behavior remains constrained by predefined environmental assumptions~\cite{grigorescu2020survey,kroemer2021review}.
A notable illustration of this is autonomous driving, wherein AI models are trained to recognize a specific set of object classes, such as pedestrians, traffic signs, and lane markings~\cite{grigorescu2020survey}. 
While such approaches are demonstrably effective within their intended domains, they remain inherently limited in scope due to their reliance on predefined taxonomies and fixed execution models. 
To provide an example, an autonomous car may navigate safely through a city, but it may fail to respond appropriately when a traffic officer overrides traffic signs or signals in a manner not represented in the vehicle's training data.
Consequently, they are unable to generalize to dynamic and open-ended environments in which robots encounter novel objects, conditions, and unknown tasks~\cite{bendale2015towards}. 

This leads to the first challenge in the field of adaptive robotics, where robots are expected to operate in environments and task settings that cannot be fully specified at design time. 
Addressing this challenge requires a shift toward natural language as a versatile medium for task specification, enabling robots to interpret and execute a virtually limitless range of complex instructions that cannot be captured by rigid, predefined taxonomies.
The inherent flexibility of natural language introduces a significant secondary challenge: such high-level intentions are frequently underspecified, leaving a wide semantic gap between a human's intent and the granular actions a robot must perform.
To illustrate this point, one may consider a scenario in which a robot is assigned the intention of requesting a coffee. 
It is conceivable that such tasks could encompass the instruction of a robot to explore an unknown environment, inspect a room, search for relevant objects, operate a machine, and retrieve a requested coffee. 
However, this instruction does not include several key elements. 
Firstly, the text provides no indication as to which objects the robot is to encounter and how to react to them. 
Secondly, the text fails to specify which areas the robot should visit first. 
Thirdly, the text does not provide a definitive timeframe within which the exploration should be deemed complete.
In conclusion, translating a simple human intention into a concrete operational plan remains a major hurdle, as the robot must autonomously infer the missing procedural details and environmental constraints.

Recent advances in generative AI offer a promising foundation for this type of intention-driven adaptation. 
LLMs have the capacity to interpret natural language instructions, infer task goals, generate action plans, and reason about task progress~\cite{belzner2023llm,ahn2022can,moon2024find,qin2024infobench}. 
Nevertheless, intention-driven robotic control also necessitates the grounding of these plans in the physical environment. 
To achieve this, VLMs provide a mechanism for deriving semantic descriptions from visual observations, enabling robots to reason about objects and spatial relations that extend beyond predefined categories~\cite{song2025avi}.
The integration of generative AI into robotic control systems introduces a fundamental tension, as the properties that render generative models attractive for open-ended adaptation also render them difficult to integrate into reliable control loops. 
It is evident that both LLMs and VLMs possess a degree of flexibility that enables them to function on instructions that are underspecified and on incomplete scene information. 
However, it is important to acknowledge that the outputs of these systems are non-deterministic and not guaranteed to be correct. 
Furthermore, these outputs are based on continuously evolving representations of the environment. 
For CAS, this necessitates a framework where generative AI is embedded into an adaptive feedback architecture without compromising reliability. 
This challenge motivates the following research question:

\begin{quote}
How can generative AI be integrated into a Collective Adaptive System for robotic control such that user intentions can be translated into reliable robot behavior under uncertainty, incomplete knowledge, and continuous environmental change?
\end{quote}

The question is addressed by means of an architectural framework for intent\-ion-dri\-ven robotic task execution in dynamic and unstructured environments. 
The architectural framework implements the MAPE-K feedback loop, meticulously decomposing the process of robotic adaptation into five distinct phases: monitoring, analysis, planning, execution, and knowledge management~\cite{arcaini2015modeling}. 
A VLM-based perception component is responsible for monitoring the environment and producing open-ended semantic descriptions of the robot's surroundings. 
A planning component, based on an LLM, is capable of interpreting user intentions and generating executable robot behavior. 
A library of robotic control functions constrains the LLM's output to valid actions, thereby enabling syntactic verification before execution. 
A workflow-monitoring and knowledge component maintains information about execution progress, previous actions, and observed scene changes, thereby supporting adaptation over long-running tasks.

Our architectural framework addresses three fundamental challenges in the domain of intention-driven adaptive robotics. 
Firstly, it mitigates the formalization gap between imprecise natural language intentions and the precise control structures required for robotic execution. 
Our framework employs an advanced LLM that translates user intentions into workflows. 
Secondly, it addresses the taxonomy gap that arises when robots encounter objects and situations that cannot be fully anticipated at design time. 
Rather than relying exclusively on a fixed object taxonomy, the VLM provides open-ended semantic grounding, which is stabilized over time through repeated observations. 
Thirdly, it addresses the temporal state problem of generative controllers by maintaining explicit execution state, thus allowing the system to distinguish between completed subtasks, failed actions, and pending objectives.

In order to investigate the aforementioned research question, a proof-of-concept implementation of the proposed architecture for a quadruped robot performing exploration tasks is presented. 
The robot receives high-level user intentions rather than explicit programs and must autonomously generate, execute, and adapt task behavior in response to its environment. 
The system is evaluated across a range of object-identification and exploration tasks and scenarios, utilizing both finite and open-ended tasks. 
Furthermore, a range of LLMs are being benchmarked to ascertain how variations in generative models impact task success, robustness, replanning behavior, and termination control.

The following contributions are made by the paper:

\begin{enumerate}
    \item The present paper proposes a MAPE-K-inspired architectural framework for intention-driven robotic control in dynamic and unstructured environments.
    \item The integration of LLMs and VLMs into a constrained adaptive feedback loop that combines semantic grounding, executable code generation, monitoring, and persistent task knowledge is demonstrated.
    \item The performance of three state-of-the-art LLMs is evaluated as a fundamental component for intention-driven robotic control in dynamic and unstructured environments, utilizing exploration tasks that necessitate spatial and temporal reasoning, executable code generation, programming capabilities, and mathematical problem-solving.
\end{enumerate}

In \Cref{section:framework-architecture} the framework architecture is presented. 
While \Cref{section:methods-and-implementation} describes the methods and implementation, \Cref{section:evaluation} shows an evaluation. 
After \Cref{section:related-work} addresses the related work, \Cref{section:conclusion} concludes our work.
\section{Framework Architecture}
\label{section:framework-architecture}
\begin{figure}
    \centering
    \includegraphics[width=0.8\linewidth]{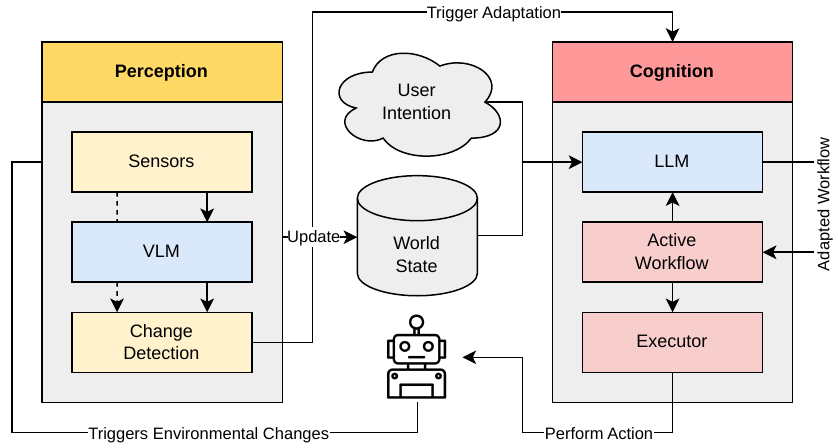}
    \caption{MAPE-K influenced framework architecture for autonomous robots resolving user intentions in dynamic and unstructured environments.}
    \label{figure:framework-architecture}
\end{figure}

The proposed framework architecture is founded on the principles of MAPE-K, a standard architectural pattern in CAS~\cite{arcaini2015modeling}. 
This loop, comprising Monitoring, Analysis, Planning, and Execution, operates over a shared knowledge base that functions as a stigmergic medium. 
As illustrated in \Cref{figure:framework-architecture}, the framework architecture is divided into a perception component, a cognition component, and a centralized world state. 
Rather than serving as a passive repository, this world state provides the environmental cues and geometric grounding necessary to drive adaptation through the environment itself.
The framework architecture is specifically engineered to address the inherent complexities of autonomous robotic adaptation by targeting three critical challenges: the taxonomy gap, formalization gap, and the temporal state problem. 

The perception component mitigates the taxonomy gap, the discrepancy between a robot's fixed internal categories and the infinite variety of the physical world, by decoupling geometric grounding from semantic identification. 
Instead of relying on closed-set classifiers, the system utilizes a VLM to perform open-ended semantic enrichment of raw sensor data. 
This information is synchronized with the world state, which functions not as a passive repository but as a stigmergic medium that provides the environmental cues necessary for cognitive reasoning. 
By grounding abstract semantic labels in precise geometric coordinates, the system ensures that the robot can interact with and reason about previously unidentified objects without requiring a predefined taxonomic library.

Bridging the formalization gap between high-level user intentions and low-level robotic control requires a rigorous approach. 
Within the cognition component of this framework architecture, processes are organized around workflows modeled as Turing-complete imperative programs. 
This structure allows the system to map ambiguous natural language intentions onto a formal sequence of actions restricted to a predefined library of high-level robotic primitives for movement control, ensuring both safety and controllability. 
The LLM acts as the central decision-making entity within this planning phase, analyzing the current world state and the active workflow to generate updated logic in response to environmental triggers. 
This transformation of intent into executable code provides a formal bridge between the non-deterministic nature of human language and the deterministic requirements of robotic  execution.

To maintain operational continuity during these adaptations, the framework architecture explicitly addresses the temporal state problem, which occurs when a system loses track of its execution progress upon receiving a new plan. The executor facilitates this by utilizing a virtual machine capable of capturing snapshots of the internal execution state, including instruction pointers and variable bindings. When a change in the environment triggers a re-planning event, the current state is preserved and provided to the LLM, enabling the synthesis of an adapted workflow that can be hot-reloaded mid-execution. This approach ensures that the robot does not reset its task upon every environmental shift, but rather adapts its behavior while maintaining full state awareness. Consequently, the entire MAPE-K loop operates as an event-driven paradigm where the system continuously perceives its surroundings, reasons about semantic and geometric changes, and executes formal, state-aware adaptations in real-time.
\section{Methods and Implementation}
\label{section:methods-and-implementation}
The realization of the framework architecture described in \Cref{section:framework-architecture} requires methods to ground environmental data and formalize the adaptation loop. 
This section details the algorithmic approach to perception, the mathematical definition of adaptation triggers, and the implementation of the perception and cognition components.

\subsection{Perception Component}
The perception layer converts raw sensor streams into the stigmergic world state. 
This process relies on geometric clustering to identify physical entities and asynchronous vision-language inference for semantic enrichment.
Even in static settings, the internal model remains highly dynamic due to the incremental nature of discovery. 
An object's full semantic and spatial profile is rarely visible from a single perspective. 
To resolve this, the component constantly fuses heterogeneous sensor data, while the robot is exploring the world: 2D LiDAR provides geometric dimensions and distances, while a VLM provides unrestricted natural language descriptions. 
This fusion allows the system to manage a shifting world state, where initial geometric hypotheses are continuously refined by new viewing angles and semantic updates.

\subsubsection{Geometric Grounding and AABB Logic}
To bridge the gap between raw, unstructured sensor data and the higher-level semantic understanding required by the LLM, we must transform LiDAR observations into stable world observations.
Therefore, physical entities are represented as Axis-Aligned Bounding Boxes (AABB) in the Cartesian plane. 
To transform unstructured LiDAR point clouds into entities within the perception layer, the system utilizes a sequential clustering algorithm that maintains spatial consistency by accounting for the decrease in ray density. For any two sequential points $p_i$ and $p_{i-1}$ located at radial distances $d_i$ and $d_{i-1}$ from the sensor, the dynamic distance threshold $\delta_{arc}$ is defined as:

\begin{equation}
\delta_{arc} = C \cdot \max(d_i, d_{i-1}) \cdot \theta_{res}
\end{equation}

where $\theta_{res}$ represents the angular resolution in radians, and $C$ is a constant multiplier acting as a tolerance buffer for sensor noise.
Points are clustered if their Euclidean distance is less than $\delta_{arc}$. 
Once a set of points $\mathcal{P} = \{p_1, \dots, p_n\}$ is assigned to a cluster, the system defines the entity as an Axis-Aligned Bounding Box whose boundaries are the minimum and maximum coordinates of the point set $\mathcal{P}$ along each global axis:
\begin{equation}
x_{min} = \min_{p \in \mathcal{P}} x_p, \quad x_{max} = \max_{p \in \mathcal{P}} x_p, \quad y_{min} = \min_{p \in \mathcal{P}} y_p, \quad y_{max} = \max_{p \in \mathcal{P}} y_p
\end{equation}
To ensure the robustness of the world state, clusters are only promoted to formal AABB entities if they exceed a density threshold $N_{min}$, defined as $1\%$ of the total LiDAR ray count. 
Temporal stability of AABB entities over time is maintained through an Exponential Moving Average (EMA) with a distance-dependent learning rate $\alpha = \exp(-d)$, requiring the world state to be updated with every LiDAR measurement as new spatial evidence is gathered. 
This continuous refinement is necessary to mitigate sensor imprecision and consolidate varying perspectives as the robot views objects from different angles.

\subsubsection{Asynchronous Semantic Enrichment}
Semantic data is integrated into the world state through a Vision-Language Model (VLM). 
When an AABB enters the camera's field of view, the system triggers an asynchronous inference request using a centered RGB image. 
The VLM estimates the object's physical footprint and provides natural language tags. 
These tags are subsequently fused into the World State by performing a spatial join between the VLM estimated AABB and the existing LiDAR derived hitboxes.
To mitigate transient hallucinations, the system maintains a running frequency distribution of tags for each entity. 
Only the top-$k$ most frequent tags are surfaced to the cognition component, ensuring that semantic grounding remains statistically stable across multiple frames.

\subsection{Event-Driven Adaptation Triggers}
The transition from monitoring to planning, the Analysis phase of the MAPE-K loop, is governed by three mathematical triggers. 
These triggers serve as gatekeepers to ensure the Large Language Model (LLM) is invoked only when environmental shifts exceed significant thresholds. 
The \textbf{New Object Trigger} is activated immediately upon the promotion of a new AABB from a raw cluster to a formal entity in the world state. 
The \textbf{Geometric Change Trigger} monitors the historical area $A$ of active AABBs. A re-planning request is issued if the current area $A_t$ satisfies the growth condition $\frac{A_t}{A_{initial}} > \gamma$, where $A_{initial}$ is the first stable area in the history buffer and $\gamma$ is a predefined growth threshold. 
This captures significant discoveries, such as the revelation of an occluded surface. 
Finally, the \textbf{Semantic Discovery Trigger} is activated if the VLM inference results in a shift where $n$ out of the $k$ most frequent tags change, indicating that the understood nature of an entity has fundamentally evolved.

\subsection{Cognition Component}
The cognition component functions as the event-driven loop mechanism performing monitoring, analyzing, planning, and execution using an LLM and the executor by incorporating an active workflow, a user intention and adaptation triggers.

A workflow is formally represented as a program written with a Turing-complete imperative programming language. For the purposes of this implementation, a workflow is represented as a Python 3 program, based on the assumption that LLMs are well-trained on publicly available data and Python 3 is one of the most popular programming languages according to the Tiobe index~\cite{index2023tiobe}.
The executor comprises a source-to-bytecode compiler utilizing the default Python 3 implementation, CPython, and a custom stack-based immutable virtual machine for executing Python 3 bytecode.

The virtual machine provides isolation and controlled execution, whilst also supporting granular hot code reloading. This enables dynamic updates to workflows without compromising runtime operation. In order to facilitate adaptation, the virtual machine creates snapshots of its internal state. These snapshots include the current instruction, variable bindings, function parameters and call stack. This facilitates seamless recovery and modification, thereby ensuring robust and fault-tolerant execution in dynamic environments.

\section{Evaluation}
\label{section:evaluation}
In order to validate the proposed framework architecture, the system is evaluated on a benchmarking suite specifically designed for quadruped robot exploration tasks. The benchmarks target the mitigation of taxonomy, formalization, and temporal state gaps by testing the robot's ability to interpret high-level exploration goals, maintain consistent state representations over time, and execute adaptive behaviors in partially observable environments. This setting facilitates an evaluation of the architecture's capacity to facilitate effective decision-making processes for embodied agents operating within complex and dynamically evolving terrain.

\subsection{Experimental Setup}
We evaluate our system using Gazebo as a simulation environment, which communicates with the aforementioned components using the Robot Operating System 2 (ROS 2). The simulated quadruped robot contains an RGB camera with a resolution of 640x480 pixels and a 2D-LiDAR with a 360\textdegree\ resolution of 512 rays.
An environmental setup is created that varies environmental visibility and task complexity to measure the system's adaptive performance.

\subsubsection{Environments}
\begin{figure}[h]
    \centering
    \includegraphics[width=0.8\textwidth]{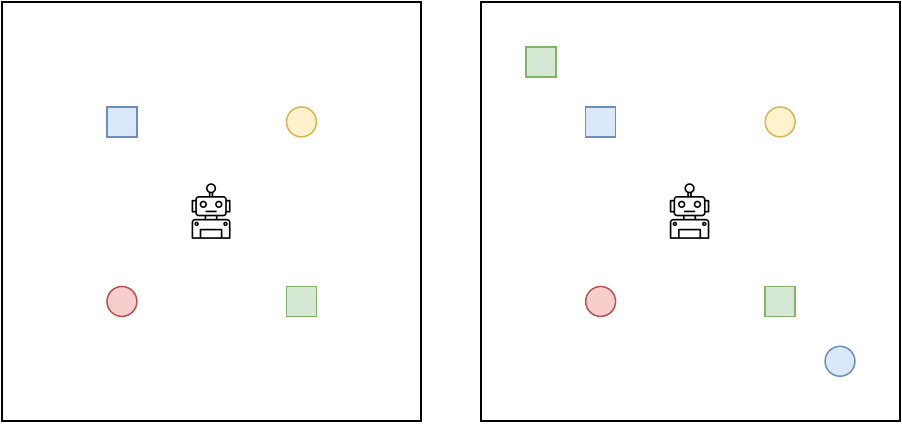}
    \caption{Evaluation environments. Left: Environment A (Full Visibility). Right: Environment B (Incremental Discovery).}
    \label{fig:evaluation_environments}
\end{figure}

As illustrated in~\autoref{fig:evaluation_environments}, our system is evaluated through two experimental environments.

\begin{itemize}
    
    \item \textbf{Environment A (Full Visibility):} The robot commences with all objects within the LiDAR’s sensing range, requiring the system to subsequently enrich these geometric detections with semantic attributes via the VLM.

    \item \textbf{Environment B (Incremental Discovery):} Significant objects are initially out of view, requiring active exploration. This environment stress-tests the new object trigger, as newly discovered entities can invalidate the current plan.

\end{itemize}

\subsubsection{Tasks}
To assess the system's adaptive capabilities and its transition from closed-world execution to open-ended discovery, we categorize the evaluation into two task types:

\paragraph{Finite Tasks}
These tasks incorporate explicit quantitative constraints to evaluate the monitor's efficacy in tracking progress and mitigating state amnesia. Specifically, the robot is assigned objectives such as "Move to two squared objects", requiring consistent state maintenance and objective fulfillment. 

\paragraph{Infinite Tasks}
These involve open-ended, categorical objectives, such as "Move to all squared objects", requiring the system to maintain a continuous discovery loop. 
Unlike finite tasks, infinite tasks lack a predefined target count. Consequently, the robot must dynamically update its execution plan whenever the perception layer identifies additional matching entities.

\paragraph{Operational Modalities}
We distinguish between goal-oriented navigation "Go to the red object" and semantic exploration "Explore until you find the red object". 
While both begin in unmapped environments, the former targets identified entities, whereas the latter mandates active search~\cite{chaplot2020object}.

\subsection{Benchmarking Suite}
We evaluate the system using a cross-model matrix to demonstrate that the reliability of the adaptive loop is a fundamental property of the architecture rather than a specific generative model. 
Our selection includes frontier models optimized for code synthesis and multimodal perception, as detailed in \autoref{tab:models}.

\begin{table}[h]
\centering
\caption{LLM and VLM configurations for the benchmarking suite.}
\label{tab:models}
\begin{tabular}{@{}llccl@{}}
\toprule
\textbf{Component} & \textbf{Model} & \textbf{Year} & \textbf{Parameters} \\ \midrule
\textbf{Cognition} & Qwen3.5-122B-A10B~\cite{qwenteam2026qwen35omnitechnicalreport} & 2026 & 122B \\
& GPT-5-nano & 2025 & undisclosed \\
 & Kimi-K2.5~\cite{kimiteam2026kimik25visualagentic} & 2026 & 1.1T \\ \midrule
\textbf{Perception} & Qwen3-VL-8B-Instruct & 2025 & 8B \\ \bottomrule
\end{tabular}
\end{table}

\subsection{Metrics}
To quantify the reliability of our system, we define a set of metrics that measure its ability to maintain logical consistency while navigating an incrementally discovered environment.

\begin{itemize}
    \item \textbf{End-to-End Success Rate (SR):} Evaluates the objective fulfillment of the high-level intention. This measures the system's ability to synthesize a plan and iteratively adapt its logic as the robot’s perception of the environment evolves.
    
    \item \textbf{Syntactic Executability (SE):} Tracks how consistently the LLM generates code that adheres to operational constraints, addressing the formalization gap between fuzzy intentions and rigorous execution.
    
    \item \textbf{Re-adaptation Frequency (RF):} Quantifies the system's reactivity by calculating the ratio of plan revisions to total trigger-induced LLM calls. The metric is defined as $RF = \frac{N_{new}}{N_{total}}$, where $N_{new}$ is the count of structural plan updates and $N_{total}$ is the total number of calls initiated by the adaptive triggers.
\end{itemize}

\subsection{Quantitative Analysis}
We evaluate the system by measuring all three metrics across both Finite and Infinite tasks, as well as across the two operational modalities. 
Each task is evaluated three times to account for variability in model outputs.

\subsection{Baseline Robustness: Smoke Tests}
Before evaluating complex navigation, we perform a series of Smoke Tests. 

\begin{table}[ht]
\centering
\caption{Evaluation Results: Smoke Tests}
\label{tab:smoke_tests}
\scriptsize
\setlength{\tabcolsep}{4pt}
\renewcommand{\arraystretch}{1.3}
\begin{tabularx}{\textwidth}{@{} c >{\raggedright\arraybackslash}X l | ccc | ccc @{}}
\toprule
 & & & \multicolumn{3}{c}{\textbf{Env. A}} & \multicolumn{3}{c}{\textbf{Env. B}} \\
\cmidrule(lr){4-6} \cmidrule(lr){7-9}
\textbf{\#} & \textbf{Task Description} & \textbf{Model} & \textbf{SR} & \textbf{SE} & \textbf{RF} & \textbf{SR} & \textbf{SE} & \textbf{RF} \\ 
\midrule
\multirow[t]{3}{*}{1} 
& \multirow[t]{3}{=}{Go to 2 -3.} 
 & Qwen3.5 & \colorbox{hardgreen}{1.00} & \colorbox{hardgreen}{1.00} & 0.39 & \colorbox{hardgreen}{1.00} & \colorbox{hardgreen}{1.00} & 0.47 \\
& 
 & GPT-5-nano & \colorbox{hardgreen}{1.00} & \colorbox{hardgreen}{1.00} & 0.58 & \colorbox{softgreen}{0.67} & \colorbox{hardgreen}{1.00} & 0.75 \\
& 
 & Kimi-K2.5 & \colorbox{hardgreen}{1.00} & \colorbox{hardgreen}{1.00} & 0.53 & \colorbox{hardgreen}{1.00} & \colorbox{hardgreen}{1.00} & 0.93 \\ 
\addlinespace
\multirow[t]{3}{*}{2} 
& \multirow[t]{3}{=}{Move to 2 -3.} 
 & Qwen3.5 & \colorbox{hardgreen}{1.00} & 0.83 & 0.57 & \colorbox{hardgreen}{1.00} & \colorbox{hardgreen}{1.00} & 0.40 \\
& 
 & GPT-5-nano & \colorbox{hardgreen}{1.00} & 0.67 & 0.50 & \colorbox{hardgreen}{1.00} & 0.83 & 0.33 \\
& 
 & Kimi-K2.5 & \colorbox{hardgreen}{1.00} & \colorbox{hardgreen}{1.00} & 1.00 & \colorbox{hardgreen}{1.00} & \colorbox{hardgreen}{1.00} & 1.00 \\
\bottomrule
\end{tabularx}
\end{table}

\autoref{tab:smoke_tests} demonstrates the system's resilience against adversarial prompts. Across all models and environments, near-perfect success rates and high syntactic executability confirm that formalization constraints effectively prevent malformed outputs.

\subsection{Task Complexity and Model Performance}
The core evaluation explores the shift from Finite Tasks to Infinite Tasks, testing the system's ability to maintain an open-ended discovery loop as the robot uncovers its environment.

\begin{table}[ht]
\centering
\caption{Evaluation Results: Finite Tasks}
\label{tab:finite_results}
\scriptsize
\setlength{\tabcolsep}{2pt}
\renewcommand{\arraystretch}{1.3} 
\begin{tabularx}{\textwidth}{@{} l c >{\raggedright\arraybackslash}X l | ccc | ccc @{}}
\toprule
& & & & \multicolumn{3}{c}{\textbf{Env. A}} & \multicolumn{3}{c}{\textbf{Env. B}} \\
\cmidrule(lr){5-7} \cmidrule(lr){8-10}
\textbf{Modality} & \textbf{\#} & \textbf{Task Description} & \textbf{Model} & \textbf{SR} & \textbf{SE} & \textbf{RF} & \textbf{SR} & \textbf{SE} & \textbf{RF} \\ 
\midrule

\multirow[t]{9}{*}{Goal-Oriented} 
& \multirow[t]{3}{*}{3} & \multirow[t]{3}{=}{Go to a red object.} 
 & Qwen3.5 & \colorbox{softgreen}{0.67} & \colorbox{hardgreen}{1.00} & 0.69 & \colorbox{softgreen}{0.67} & 0.89 & 0.87 \\
& & & GPT-5-nano & \colorbox{softyellow}{0.33} & 0.73 & 0.57 & \colorbox{softyellow}{0.33} & 0.83 & 0.15 \\
& & & Kimi-K2.5 & \colorbox{softgreen}{0.67} & \colorbox{hardgreen}{1.00} & 0.74 & \colorbox{hardred}{0.00} & \colorbox{hardgreen}{1.00} & 1.00 \\ \addlinespace
 
& \multirow[t]{3}{*}{4} & \multirow[t]{3}{=}{Go to two squared objects.} 
 & Qwen3.5 & \colorbox{hardgreen}{1.00} & 0.97 & 0.74 & \colorbox{softyellow}{0.33} & 0.98 & 0.92 \\
& & & GPT-5-nano & \colorbox{softyellow}{0.33} & \colorbox{hardgreen}{1.00} & 0.57 & \colorbox{softgreen}{0.67} & 0.54 & 0.50 \\
& & & Kimi-K2.5 & \colorbox{hardgreen}{1.00} & \colorbox{hardgreen}{1.00} & 0.75 & \colorbox{hardgreen}{1.00} & \colorbox{hardgreen}{1.00} & 0.83 \\ \addlinespace
 
& \multirow[t]{3}{*}{5} & \multirow[t]{3}{=}{Go to two squared objects and tell me their colors.} 
 & Qwen3.5 & \colorbox{softgreen}{0.67} & 0.81 & 0.95 & \colorbox{hardgreen}{1.00} & \colorbox{hardgreen}{1.00} & 0.89 \\
& & & GPT-5-nano & \colorbox{softyellow}{0.33} & 0.66 & 0.96 & \colorbox{softgreen}{0.67} & 0.56 & 0.68 \\
& & & Kimi-K2.5 & \colorbox{softgreen}{0.67} & \colorbox{hardgreen}{1.00} & 1.00 & \colorbox{hardgreen}{1.00} & \colorbox{hardgreen}{1.00} & 0.91 \\ \midrule

\multirow[t]{9}{*}{Semantic Exp.} 
& \multirow[t]{3}{*}{6} & \multirow[t]{3}{=}{Explore until you find a red object.} 
 & Qwen3.5 & \colorbox{hardgreen}{1.00} & 0.95 & 0.79 & \colorbox{hardgreen}{1.00} & 0.97 & 0.84 \\
& & & GPT-5-nano & \colorbox{softgreen}{0.67} & 0.77 & 0.52 & \colorbox{softgreen}{0.67} & 0.78 & 0.79 \\
& & & Kimi-K2.5 & \colorbox{hardgreen}{1.00} & \colorbox{hardgreen}{1.00} & 0.91 & \colorbox{hardgreen}{1.00} & \colorbox{hardgreen}{1.00} & 0.70 \\ \addlinespace
 
& \multirow[t]{3}{*}{7} & \multirow[t]{3}{=}{Explore until you find two squared objects.} 
 & Qwen3.5 & \colorbox{softgreen}{0.67} & \colorbox{hardgreen}{1.00} & 0.78 & \colorbox{hardgreen}{1.00} & \colorbox{hardgreen}{1.00} & 0.89 \\
& & & GPT-5-nano & \colorbox{softyellow}{0.33} & 0.94 & 0.55 & \colorbox{softgreen}{0.67} & 0.63 & 0.65 \\
& & & Kimi-K2.5 & \colorbox{hardgreen}{1.00} & \colorbox{hardgreen}{1.00} & 0.70 & \colorbox{hardgreen}{1.00} & \colorbox{hardgreen}{1.00} & 0.84 \\ \addlinespace
 
& \multirow[t]{3}{*}{8} & \multirow[t]{3}{=}{Explore until you find two squared objects and tell me their colors.} 
 & Qwen3.5 & \colorbox{hardgreen}{1.00} & 0.93 & 0.87 & \colorbox{hardgreen}{1.00} & 0.97 & 0.88 \\
& & & GPT-5-nano & \colorbox{hardred}{0.00} & 0.55 & 0.49 & \colorbox{hardred}{0.00} & 0.68 & 0.81 \\
& & & Kimi-K2.5 & \colorbox{hardgreen}{1.00} & \colorbox{hardgreen}{1.00} & 0.93 & \colorbox{hardgreen}{1.00} & \colorbox{hardgreen}{1.00} & 0.91 \\
\bottomrule
\end{tabularx}
\end{table}

The results in \autoref{tab:finite_results} show that the proposed adaptive loop reliably preserves task progress and logical consistency in both environments. 
Across all configurations, the open-source models (Qwen3.5 and Kimi-K2.5) achieve higher success rates, especially in Environment B, indicating stronger robustness to dynamic observation updates and newly discovered objects. 
GPT-5-nano, by contrast, performs worse on tasks requiring counting and multi-step state tracking, suggesting sensitivity to the temporal state gap.
Failures were concentrated in object-referenced and exploration-based tasks, pointing to systematic weaknesses in grounding, memory, and output compliance. In Tasks 3 to 5, several models struggled sometimes to correctly map linguistic object descriptions to perceived entities. 
GPT-5-nano often focused on incorrect targets, hallucinated colors, or violated the response format. 
Kimi-K2.5 showed similar grounding failures and sometimes falsely concluded that no red objects were present under incremental discovery. 
Qwen3.5 was occasionally distracted by non-target objects.
The results demonstrate high system reliability, with successful task completion achieved in the vast majority of cases. 
Residual failures are attributable not to low-level action execution, but to specific challenges in semantic grounding, constraint adherence, and the maintenance of coherent internal state representations during sequential exploration.

\begin{table}[ht]
\centering
\caption{Evaluation Results: Infinite Tasks}
\label{tab:infinite_results}
\scriptsize
\setlength{\tabcolsep}{2pt}
\renewcommand{\arraystretch}{1.3}
\begin{tabularx}{\textwidth}{@{} l c >{\raggedright\arraybackslash}X l | ccc | ccc @{}}
\toprule
& & & & \multicolumn{3}{c}{\textbf{Env. A}} & \multicolumn{3}{c}{\textbf{Env. B}} \\
\cmidrule(lr){5-7} \cmidrule(lr){8-10}
\textbf{Modality} & \textbf{\#} & \textbf{Task Description} & \textbf{Model} & \textbf{SR} & \textbf{SE} & \textbf{RF} & \textbf{SR} & \textbf{SE} & \textbf{RF} \\ 
\midrule

\multirow[t]{3}{*}{Goal-Oriented} & \multirow[t]{3}{*}{9} & \multirow[t]{3}{=}{Go to all cylindric objects.} 
 & Qwen3.5 & \colorbox{softgreen}{0.67} & 0.93 & 0.93 & \colorbox{hardred}{0.00} & \colorbox{hardgreen}{1.00} & 0.93 \\
 & & & GPT-5-nano & \colorbox{softyellow}{0.33} & 0.87 & 0.70 & \colorbox{hardred}{0.00} & \colorbox{hardgreen}{1.00} & 0.57 \\
 & & & Kimi-K2.5 & \colorbox{hardred}{0.00} & \colorbox{hardgreen}{1.00} & 0.83 & \colorbox{hardgreen}{1.00} & 0.96 & 0.69 \\
\midrule

\multirow[t]{6}{*}{Semantic Exp.} & \multirow[t]{3}{*}{10} & \multirow[t]{3}{=}{Explore until you find all cylindric objects.} 
 & Qwen3.5 & \colorbox{hardgreen}{1.00} & \colorbox{hardgreen}{1.00} & 0.54 & \colorbox{hardgreen}{1.00} & 0.88 & 0.78 \\
 & & & GPT-5-nano & \colorbox{softgreen}{0.67} & 0.80 & 0.56 & \colorbox{softyellow}{0.33} & 0.71 & 0.41 \\
 & & & Kimi-K2.5 & \colorbox{softgreen}{0.67} & \colorbox{hardgreen}{1.00} & 0.72 & 0.67 & \colorbox{hardgreen}{1.00} & 0.69 \\ \addlinespace
 
 & \multirow[t]{3}{*}{11} & \multirow[t]{3}{=}{Explore until you find all green squared objects.} 
 & Qwen3.5 & \colorbox{softgreen}{0.67} & \colorbox{hardgreen}{1.00} & 0.44 & \colorbox{hardgreen}{1.00} & 0.92 & 0.71 \\
 & & & GPT-5-nano & \colorbox{hardred}{0.00} & 0.89 & 0.87 & \colorbox{hardred}{0.00} & 0.67 & 0.83 \\
 & & & Kimi-K2.5 & \colorbox{softgreen}{0.67} & \colorbox{hardgreen}{1.00} & 0.73 & \colorbox{softgreen}{0.67} & \colorbox{hardgreen}{1.00} & 0.77 \\
\bottomrule
\end{tabularx}
\end{table}

\begin{figure}
    \centering
    \includegraphics[width=\linewidth]{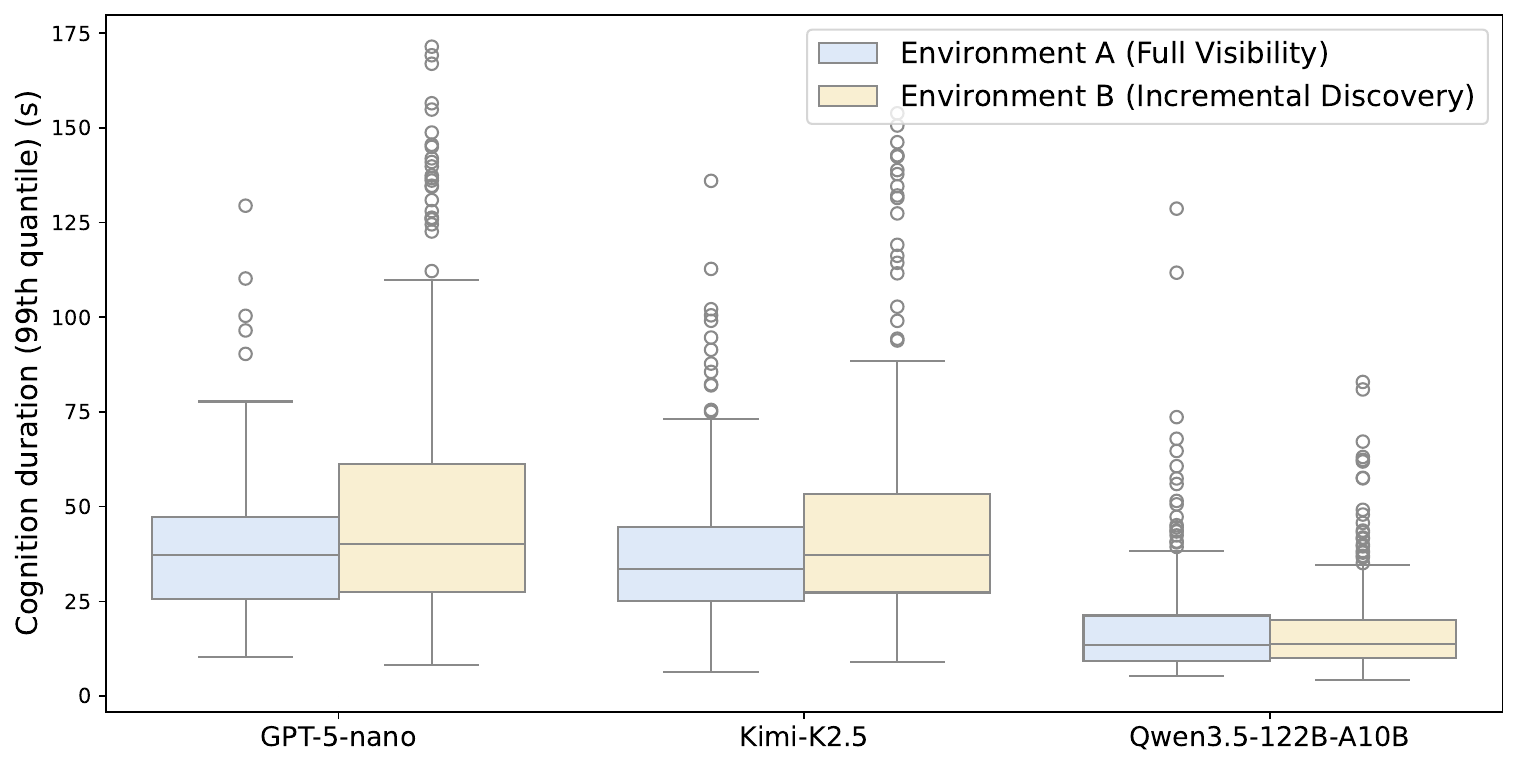}
    \caption{Cognition durations per model and environment}
    \label{fig:cognition_durations}
\end{figure}

\autoref{tab:infinite_results} shows the system’s performance on open-ended tasks under continuous environment expansion. 
Compared with finite tasks, Success Rates are more variable, with the strongest degradation in Goal-Oriented settings, especially in Environment B, where incomplete coverage of newly discovered objects becomes a key challenge. 
Nevertheless, every task was solved successfully in at least 2 out of 3 cases by at least one model, showing that none was inherently infeasible or due to luck. 
However, strong aggregate differences remain: Qwen3.5 and Kimi-K2.5 each achieve an overall success rate of about 82\%, clearly outperforming GPT-5-nano at about 46\%.
More specific, semantic exploration produces more consistent results overall, although GPT-5-nano still lags behind in reliability.
In Tasks 9 to 11, failures reveal limitations in completeness checking, category-level reasoning, and termination control. 
GPT-5-nano frequently failed to cover relevant targets, stopped prematurely, or concluded that exploration was complete despite insufficient evidence. 
Kimi-K2.5 and Qwen3.5 showed similar but less severe issues, including partial coverage, revisiting previously explored objects, and terminating despite remaining uncertainty or unidentified objects. 
Overall, these results suggest that exhaustive search over a target class is challenging primarily because the models struggle to estimate environmental completeness. 
Beyond these qualitative nuances, the computational efficiency of these models becomes clear when analyzing cognition latency.
As shown in \autoref{fig:cognition_durations}, GPT-5-nano also performs worse than both open-source models when adaptive cognition duration is considered. 
In conclusion, the Qwen3.5 model is the most robust, with a Success Rate that is equivalent to that of the Kimi-K2.5 model. It also has the shortest inference time and the fastest adaptability, despite having the smallest parameter count. While both models utilize a Mixture-of-Experts architecture, Qwen3.5 activates significantly fewer parameters per token than Kimi-K2.5, which is responsible for its lower inference latency despite comparable success rates.
It is evident that the substantial inference time per query, in conjunction with the necessity for manual evaluation of model outputs, necessitates the implementation of a maximum of three attempts per model. While this restricts the statistical robustness of the findings, it is a necessary trade-off given the computational and manual nature of the evaluation protocol.

\section{Related Work}
\label{section:related-work}
The engineering of Collective Adaptive Systems and the integration of generative AI into autonomous control loops represent a critical convergence of formal methods and adaptive robotics. 
This section reviews the foundations of methodologies for bridging the taxonomy, formalization, and temporal state gaps.

Collective Adaptive Systems, often referred to as ensembles, consist of heterogeneous components that adapt to open-ended environments through local interactions \cite{wirsing2023rigorous}. 
A foundational paradigm for managing these interactions is attribute-based communication, which allows components to interact based on runtime properties and roles rather than fixed identifiers \cite{alrahman2017abc}.
In the domain of robotics, the Protease 2.0 framework provides a parameterizable algorithmic pattern for swarm formation flight, utilizing the ATGC (Aggregation, Termination, Grouping, Calculation) pattern to enable decentralized self-organization \cite{kosak2024approach}.
Research into swarm robustness further utilizes statistical model checking to quantify the resilience of consensus-reaching behaviors in the presence of disruptive individuals \cite{klein2024consensus}.

The taxonomy gap, the inherent limitation of static command libraries in unpredictable worlds, is increasingly addressed by leveraging Vision-Language Models for open-ended scene understanding. 
Modern Vision-Language-Action (VLA) architectures, with VLA architectures initially pioneered by Google with the RT-2 model \cite{zitkovich2023rt}, such as the Avi model, propose reframing action generation as a problem of 3D volumetric inference over point clouds rather than 2D policy learning, providing robust spatial grounding \cite{song2025avi}. This includes models such as 3D-VLA \cite{zhen20243d} and Any3D-VLA \cite{fan2026any3d}, which enable three-dimensional interaction. A further step in this direction is done by the V-jepa 2 model, which enables understanding, prediction, and planning based on a video stream \cite{assran2025v}.

The formalization gap denotes the disconnect between imprecise natural language intentions and rigorous robotic execution. AISoLA research identifies LLM-assisted software engineering as a paradigm shift where the model serves as a development expert to resolve specification ambiguities and generate implementation artifacts~\cite{belzner2023llm}. 
To ensure correctness, methodologies such as test-based refinement utilize property-based testing (e.g., ScalaCheck) to iteratively validate the semantic adherence of synthesized code~\cite{aichernig2023ai}. 
Furthermore, domain-specific modeling techniques integrate structured prompt templates to constrain LLM outputs within verifiable boundaries~\cite{steffen2023modeling}.

Autonomous systems encounter a temporal state problem due to the lack of structured memory in generative controllers. Code-as-Monitor frameworks mitigate this by treating visual programming as a reactive, proactive failure-detection mechanism~\cite{zhou2025codemonitor}.
For collective systems, white-box validation techniques have been developed that integrate statistical model checking with process mining to graphically explain observed behaviors and identify unexpected deadlocks or synchronization errors during mission execution~\cite{casaluce2024whitebox}.

\section{Conclusion}
\label{section:conclusion}
This work introduces a MAPE-K-inspired framework architecture that enables autonomous robots to perform tasks in unknown and unstructured environments based on high-level user intentions rather than explicit programming.
It integrates a dual-AI system consisting of an LLM for code-based planning and a VLM for distilled semantic perception, which collectively address the formalization, taxonomy, and temporal state gaps that arise when using natural language as task description.

The evaluation of a quadruped robot in the context of exploration tasks has demonstrated that the architecture ensures syntactic reliability to a high degree and exhibits robust performance in the presence of finite tasks.
Nevertheless, infinite and open-ended tasks reveal unresolved challenges in long-horizon reasoning, state awareness, and completeness checking. 
Specifically, while open-source frontier models exhibited superior performance in maintaining execution context, performance degradation in open-ended tasks was primarily attributed to cognitive limits inherent in exhaustive search, as opposed to invalid control commands or execution errors.

Subsequent research will address the temporal state problem with greater rigor. 
The intention is to move beyond simple execution pointers by implementing formal state-machine verification, where the LLM-generated plan is dynamically mapped to a state-transition system to ensure consistency across multi-step objectives. 
Finally, the exploration of multi-agent coordination will be conducted, with the objective of enabling the system to scale its collective semantic understanding across distributed robotic platforms operating in large, unmapped environments.

%
%
%
\bibliographystyle{splncs04}
\bibliography{bib}

\end{document}